\pdfoutput=1

\documentclass[11pt]{article}

\usepackage[]{acl}

\usepackage[title]{appendix}
\usepackage{times}
\usepackage{latexsym}
\usepackage{hyperref}
\usepackage[T1]{fontenc}
\usepackage[utf8]{inputenc}
\usepackage{microtype}
\usepackage{svg}
\usepackage{caption}
\usepackage{subcaption}
\usepackage{bm}
\usepackage{url}
\usepackage{lipsum}
\usepackage{multirow}
\usepackage{graphicx}
\usepackage{amsmath}

\title{Revisiting Checkpoint Averaging for Neural Machine Translation}

\author{
Yingbo Gao \qquad Christian Herold \quad Zijian Yang \qquad Hermann Ney \\
Human Language Technology and Pattern Recognition Group \\
Computer Science Department\\
RWTH Aachen University \\
D-52056 Aachen, Germany \\
{\tt \{ygao|herold|zyang|ney\}@cs.rwth-aachen.de}
}

\begin{document}

\maketitle

\begin{abstract}
Checkpoint averaging is a simple and effective method to boost the performance of converged neural machine translation models.
The calculation is cheap to perform and the fact that the translation improvement almost comes for free, makes it widely adopted in neural machine translation research.
Despite the popularity, the method itself simply takes the mean of the model parameters from several checkpoints, the selection of which is mostly based on empirical recipes without many justifications.
In this work, we revisit the concept of checkpoint averaging and consider several extensions.
Specifically, we experiment with ideas such as using different checkpoint selection strategies, calculating weighted average instead of simple mean, making use of gradient information and fine-tuning the interpolation weights on development data.
Our results confirm the necessity of applying checkpoint averaging for optimal performance, but also suggest that the landscape between the converged checkpoints is rather flat and not much further improvement compared to simple averaging is to be obtained.
\end{abstract}

\section{Introduction}

Checkpoint averaging is a simple method to improve model performance at low computational cost.
The procedure is straightforward: select some model checkpoints, average the model parameters, and obtain a better model.
Because of its simplicity and effectiveness, it is widely used in neural machine translation (NMT), e.g. in the original Transformer paper \cite{vasvani2017attention}, in systems participating in public machine translation (MT) evaluations such as Conference on Machine Translation (WMT) \cite{WMT:2021} and the International Conference on Spoken Language Translation (IWSLT) \cite{anastasopoulos-etal-2022-findings}: \citet{WMT:2021, erdmann-gwinnup-anderson:2021:WMT, li-EtAl:2021:WMT1, subramanian-EtAl:2021:WMT, tran-EtAl:2021:WMT, wang-EtAl:2021:WMT1, wei-EtAl:2021:WMT, di-gangi-etal-2019-data, li-etal-2022-hw}, and in numerous MT research papers \cite{junczys-dowmunt-etal-2016-neural, shaw-etal-2018-self, liu2018comparable, zhao2019muse, kim2021scalable}.
Apart from NMT, checkpoint averaging also finds applications in Transformer-based automatic speech recognition models \cite{Karita2019ACS, dong2018speech, higuchi2020maskctc, tian2020spike, Wang2020TransformerBasedAM}.
Despite the popularity of the method, the recipes in each work are rather empirical and do not differ much except in how many and exactly which checkpoints are averaged.

In this work, we revisit the concept of checkpoint averaging and consider several extensions.
We examine the straightforward hyperparameters like the number of checkpoints to average, the checkpoint selection strategy and the mean calculation itself.
Because the gradient information is often available at the time of checkpointing, we also explore the idea of using this piece of information.
Additionally, we experiment with the idea of fine-tuning the interpolation weights of the checkpoints on development data.
As reported in countless works, we confirm that the translation performance improvement can be robustly obtained with checkpoint averaging.
However, our results suggest that the landscape between the converged checkpoints is rather flat, and it is hard to squeeze out further performance improvements with advanced tricks.

\begin{figure*}
     \centering
     \begin{subfigure}[b]{0.32\textwidth}
         \centering
         \includegraphics[width=\textwidth]{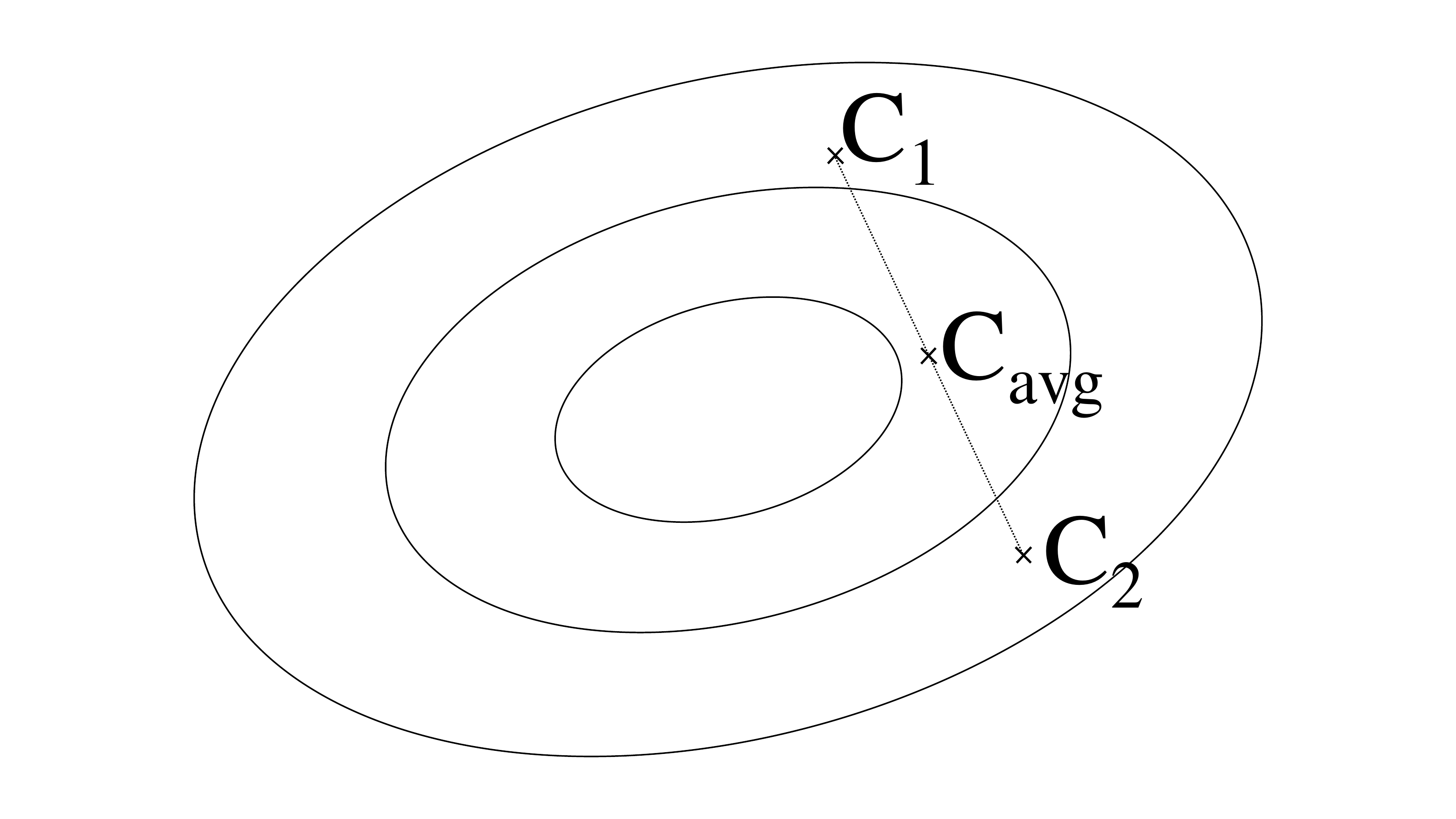}
         \caption{vanilla}
         \label{fig:1}
     \end{subfigure}
     \hfill
     \begin{subfigure}[b]{0.32\textwidth}
         \centering
         \includegraphics[width=\textwidth]{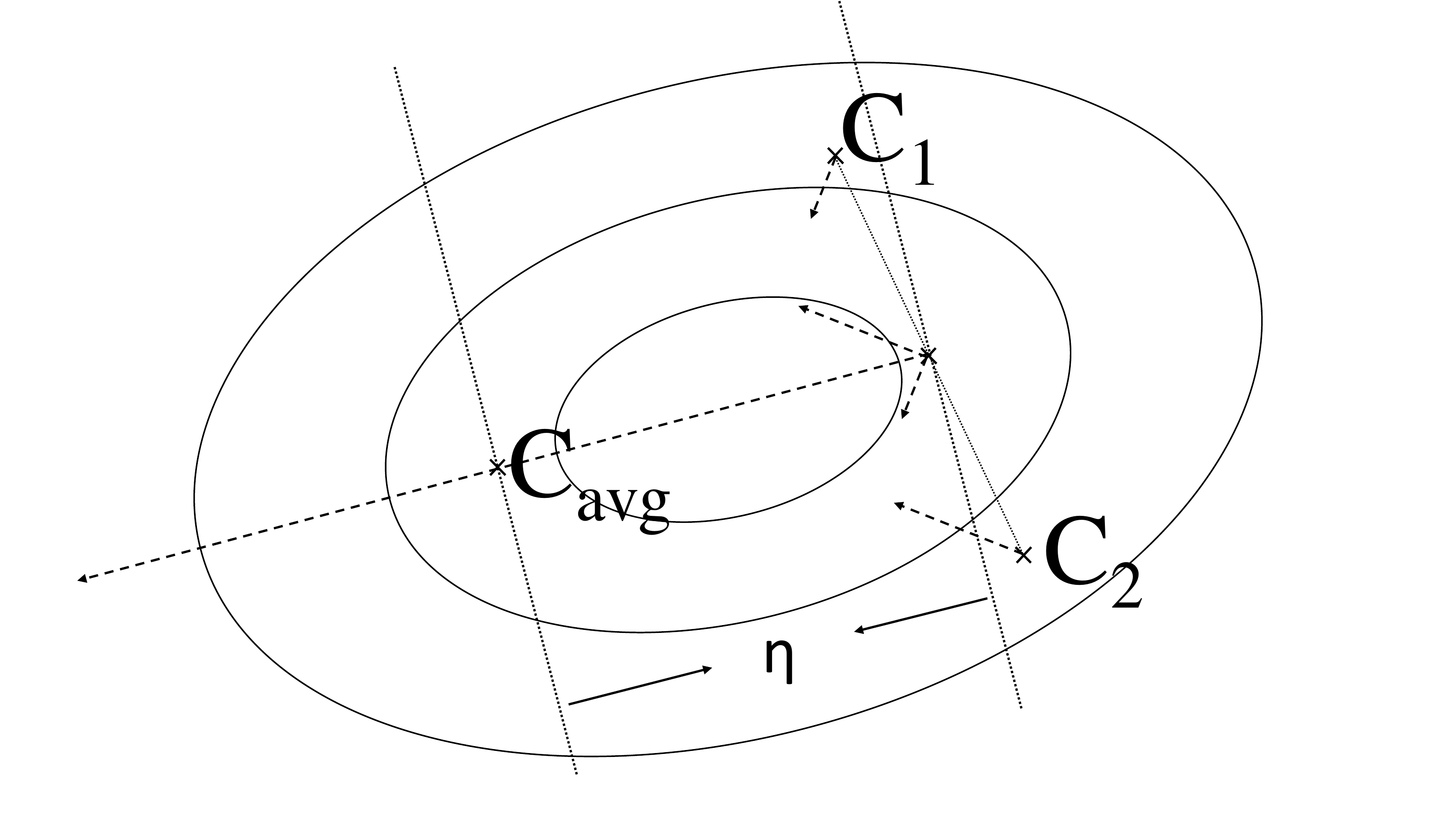}
         \caption{using gradient information}
         \label{fig:2}
     \end{subfigure}
     \hfill
     \begin{subfigure}[b]{0.32\textwidth}
         \centering
         \includegraphics[width=\textwidth]{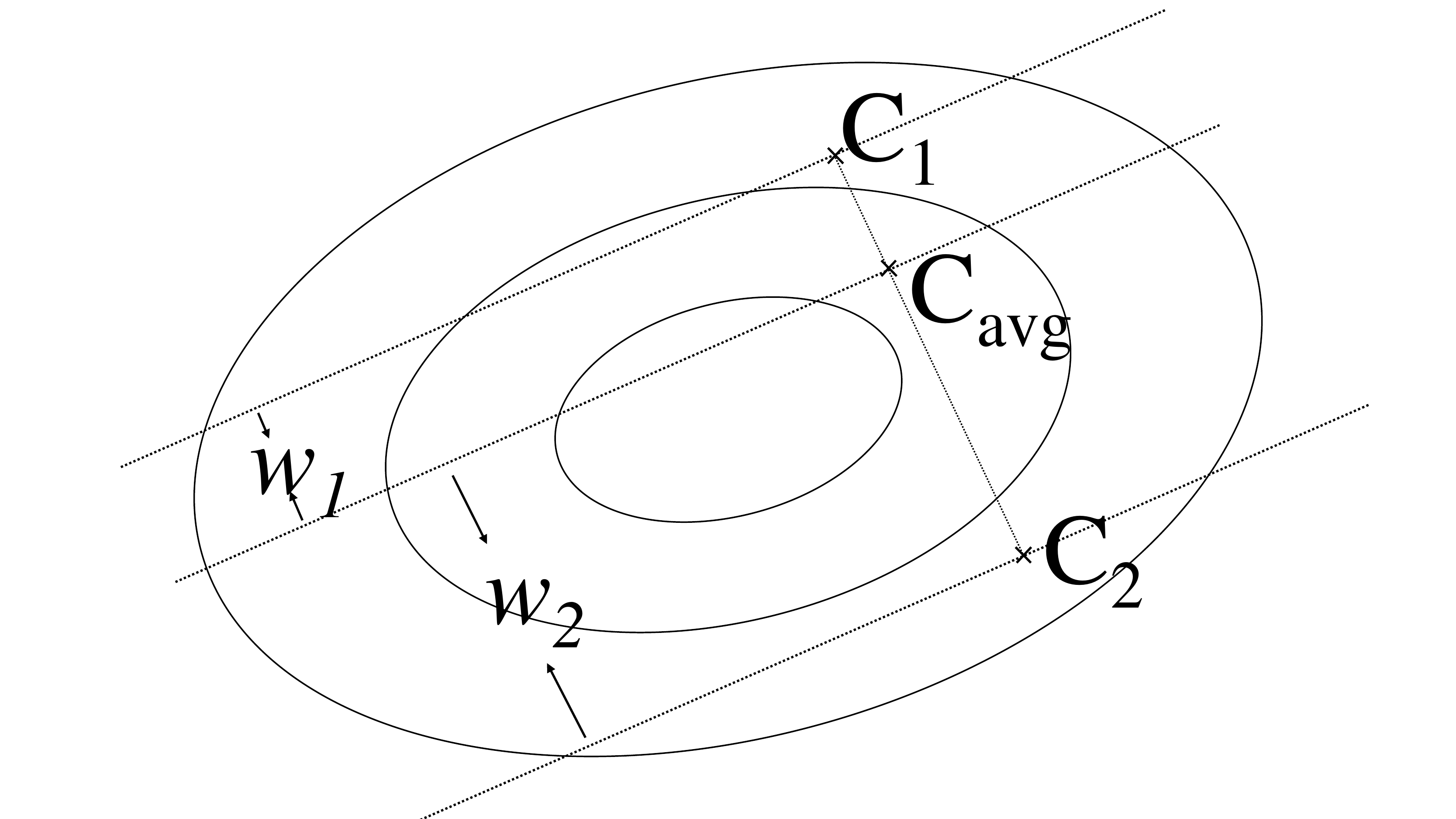}
         \caption{optimized on development data}
         \label{fig:3}
     \end{subfigure}
        \caption{
        An illustration of checkpoint averaging and our extensions.
        The isocontour plot illustrates some imaginary loss surface.
        C1 and C2 are model parameters from two checkpoints.
        C$_\mathrm{avg}$ denotes the averaged parameters.
        In (a), the mean of the C1 and C2 is taken.
        In (b), the dashed arrows refer to the gradients (could also include the momentum terms) stored in the checkpoints, and a further step (with step size $\eta$) is taken.
        In (c), a NN is parametrized with the interpolation weights $w_1$ and $w_2$, and the weights are learned on the development data.}
        \label{fig.123}
\end{figure*}

\section{Related Work}

The idea of combining multiple models for more stable and potentially better prediction is not new in statistical learning \cite{dietterich2000ensemble, dong2020survey}. In NMT, ensembling, more specifically, ensembling systems with different architectures is shown to be helpful \cite{stahlberg-etal-2019-cued, rosendahl2019rwth, zhang-van-genabith-2019-dfki}. In contrary, checkpoint averaging uses checkpoints from the same training run with the same neural network (NN) architecture. Compared to ensembling, checkpoint averaging is cheaper to calculate and does not require one to store and query multiple models at test time. The distinction can also be made from the perspective of the interpolation space, i.e. model parameter space for checkpoint averaging, and posterior probability space for ensembling. As a trade-off, the performance boost from checkpoint averaging is typically smaller than ensembling \cite{liu2018comparable}.

In the literature,
\citet{chen2017checkpoint} study the use of checkpoints from the same training run for ensembling;
\citet{smith2017cyclical} proposes cyclic learning rate schedules to improve accuracy and convergence;
\citet{gao2017snapshot} propose to use a cyclic learning rate to obtain snapshots of the same model during training and ensemble them in the probability space;
\citet{izmailov2018averaging} perform model parameter averaging on-the-fly during training and argue for better generalization in this way;
\citet{popel2018training} discuss empirical findings related to checkpoint averaging for NMT;
\citet{Zhang2020PushingTL} and \citet{Karita2021ACS} maintain an exponential moving average during model training;
\citet{wang2021boost} propose a boosting algorithm and ensemble checkpoints in the probability space;
\citet{matena2021merging} exploit the Fisher information matrix to calculate weighted average of model parameters.
Here, we are interested in the interpolation happening in the model parameter space, and therefore restrain ourselves from further discussing topics like ensembling or continuing training on the development data.

\section{Methodology}

In this section, we discuss extensions to checkpoint averaging considered in this work.
An intuitive illustration is shown in Fig.\ref{fig.123}.

\subsection{Extending Vanilla Checkpoint Averaging}

The vanilla checkpointing is straightforward and can be expressed as in Eq.\ref{eq.vani.}. Here, $\bm{\theta}$ denotes the model parameters and $\hat{\bm{\theta}}$ is the averaged parameters. $k$ is a running index in number of checkpoints $K$, and $\mathcal{S}$, where $|S|=K$, is a set of checkpoint indices selected by some specific strategy, e.g. top-$K$ or last-$K$.
In the vanilla case, $w_k = \frac{1}{K}$, i.e. uniform weights are used.

\begin{equation}
    \label{eq.vani.}\hat{\bm{\theta}} = \sum_{k \in \mathcal{S}} w_k \bm{\theta}_k
\end{equation}

As shown in Eq.\ref{eq.extd.}, we further consider non-uniform weights and propose to use softmax-normalized logarithm of development set perplexities (\textsc{DevPpl}) with temperature $\tau$ as interpolation weights.
We define $w$ in this way such that it is in the probability space.

\begin{equation}
  \begin{gathered}
    \label{eq.extd.}
    w_k = \frac {\exp(- \tau \log \mathrm{\textsc{DevPpl}}_k)} {\sum_{k' \in \mathcal{S}} \exp (- \tau \log \mathrm{\textsc{DevPpl}}_{k'})}
  \end{gathered}
\end{equation}

\subsection{Making Use of Gradient Information}

Nowadays, NMT models are commonly trained with stated optimizers like Adam \cite{kingma2015adam}.
To provide the "continue-training" utility, the gradients of the most recent batch are therefore also saved.
Shown in Eq.\ref{eq.grad.}, we can therefore take a further step in the parameter space during checkpoint averaging to make use of this information.
Here, $\eta$ is the step size and $\frac{1}{K}\sum_{k \in \mathcal{S}} \nabla_{\bm{\theta}} L(\bm{\theta}_k)$ is the mean of the gradients stored in the checkpoints. 

\begin{equation}
  \begin{gathered}
    \label{eq.grad.}\hat{\bm{\theta}} = \sum_{k \in \mathcal{S}} w_k\bm{\theta}_k - \eta \frac{1}{K}\sum_{k \in \mathcal{S}} \nabla_{\bm{\theta}} L(\bm{\theta}_k)
  \end{gathered}
\end{equation}

\subsection{Optimization on Development Data}\label{sec.opti.}

In addition to using \textsc{DevPpl}, one can optimize the interpolation weights directly on the development data.
Specifically, to ensure normalization, we re-parameterize the model with the logits $g_k$ in a softmax function, initialized at zero and updated via one-step gradient descent, with step size $\eta$, on development data to avoid overfitting.
As shown in Eq.\ref{eq.opti.}, $w_k$ is the normalized interpolation weights.
Note that we refrain from updating the raw model parameters $\bm{\theta}_k$ from each checkpoint but only update the logits $g_k$.
Here, $L$ refers to the cross entropy loss of the re-parametrized NN on the development data.

\begin{equation}
  \begin{gathered}
    \label{eq.opti.}
    w_k = \frac{\exp g_k}{\sum_{k' \in \mathcal{S}} \exp g_{k'}}\\
    g_{k, 0} = 0,\;\;g_{k, 1} = - \eta \nabla_{g_k} L(g_{k, 0}; \bm{\theta}_1, ..., \bm{\theta}_K)
  \end{gathered}
\end{equation}

\section{Experiments}\label{sec.exp.}

We re-implement Transformer \cite{vasvani2017attention} using PyTorch \cite{paszke2019pytorch} and experiment on IWSLT14 German-, Russian-, and Spanish-to-English (de-en, ru-en, es-en), and WMT16 English-to-Romanian, WMT14 English-to-German, WMT19 Chinese-to-English (en-ro, en-de, zh-en) datasets.
Due to limited length, we only present representative results on de-en in this section.
Results on other language pairs can be found in the appendix and the trends are similar to that reported in this section.
Note that, in the experiments below, the test \textsc{Bleu} scores are under consideration.
However, we argue that it is not critical because checkpoint averaging is a vetted trick to boost system performance and our goal is to better understand the parameter space and not to obtain "the state-of-the-art" in some public scoreboard.

\begin{figure}[ht]
    \centering
    \includegraphics[width=0.48\textwidth]{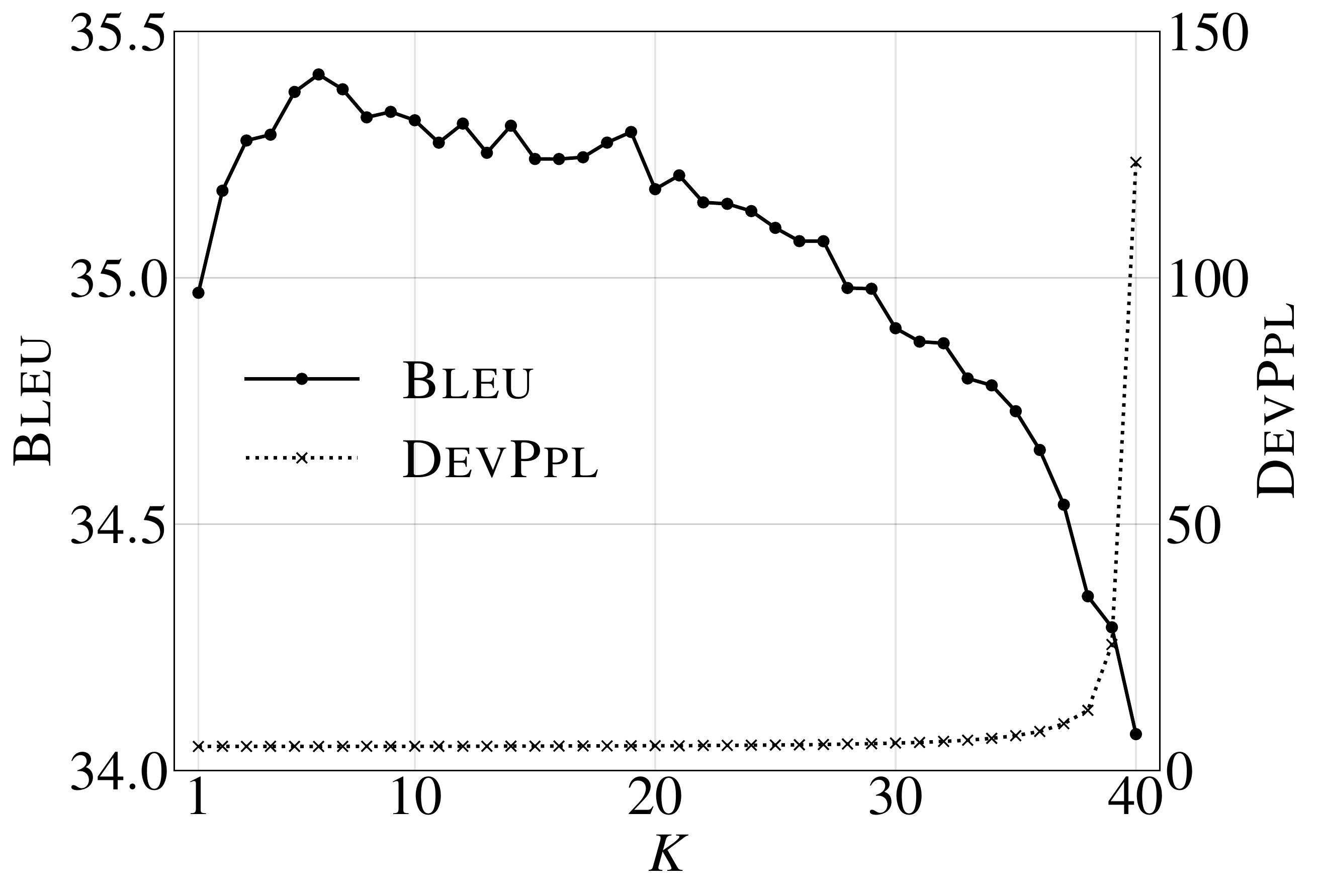}    \caption{Last-$K$ simple mean on de-en.}
    \label{fig.de-en.lastK}
\end{figure}

In Fig.\ref{fig.de-en.lastK}, we plot the \textsc{Bleu} \cite{papineni-etal-2002-bleu} scores versus increasing $K$, where the previous $K$ checkpoints starting from the best checkpoint (in terms of \textsc{DevPpl}) are selected.
As can be seen, initial \textsc{Bleu} improvements are obtained but as worse and worse checkpoints are included, the \textsc{Bleu} score drops as expected.
% comment Christian: BLEU/DEVPPL axis labels have different size especially between fig 2 & 3

\begin{figure}[ht]
    \centering
    \includegraphics[width=0.42\textwidth]{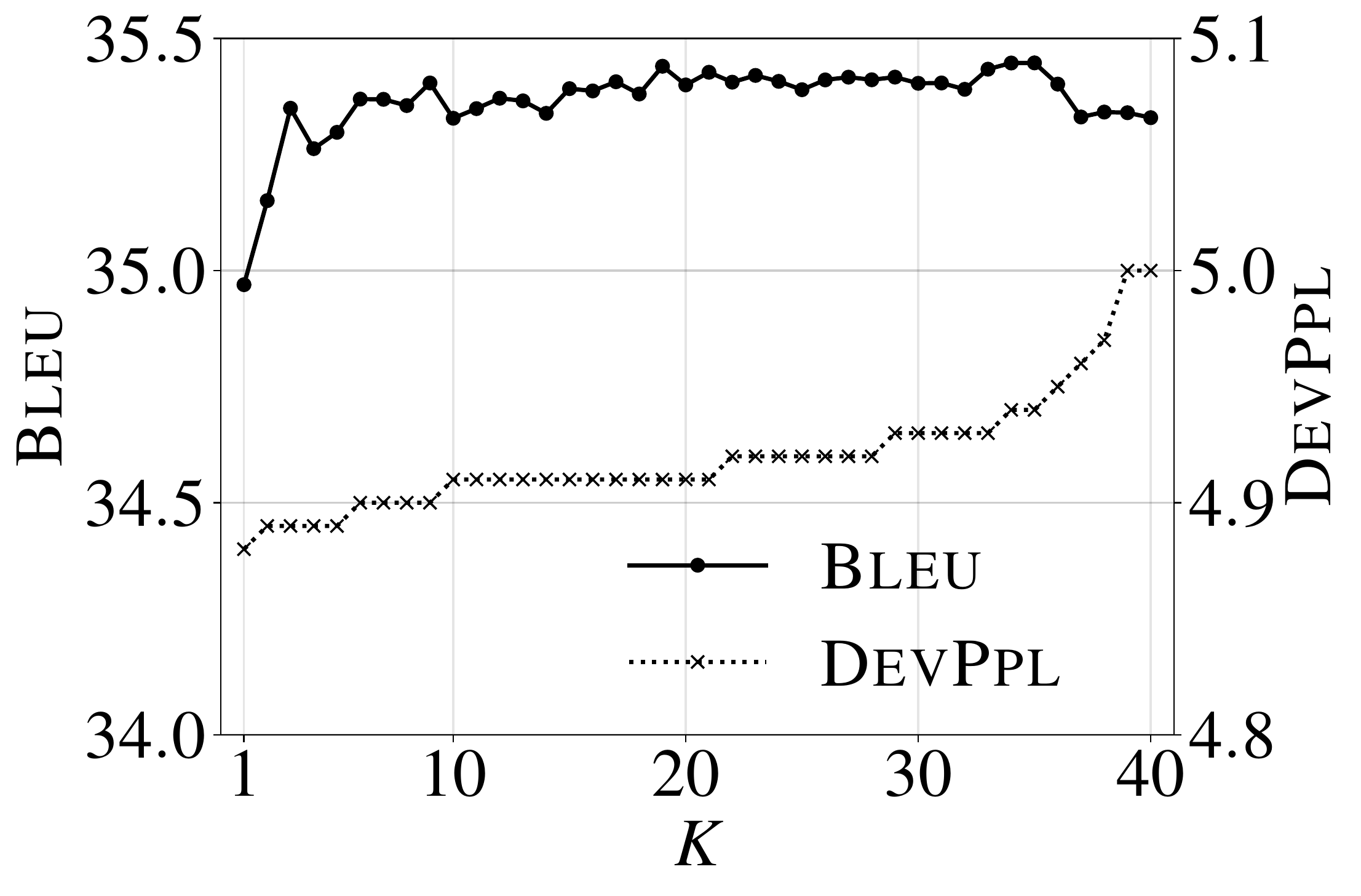}
    \caption{Top-$K$ simple mean on de-en.}
    \label{fig.de-en.topK}
\end{figure}

In Fig.\ref{fig.de-en.topK}, ranking all checkpoints by their \textsc{DevPpl}, the top-$K$ checkpoints are selected for averaging.
Notice that up to $K=40$, the \textsc{DevPpl} is still around 5, whereas in the last-$K$ case, significantly worse checkpoints (the early checkpoints) are already included in the interpolation.
It can be seen that the final \textsc{Bleu} score is much less sensitive to the choice of $K$ in this case.
Of course the final performance also relies on the checkpointing settings (e.g. the checkpointing frequency) but it is clear from the comparison that one should prefer to include checkpoints with better \textsc{DevPpl}.

\begin{figure}[ht]
    \centering
    \includegraphics[width=0.38\textwidth]{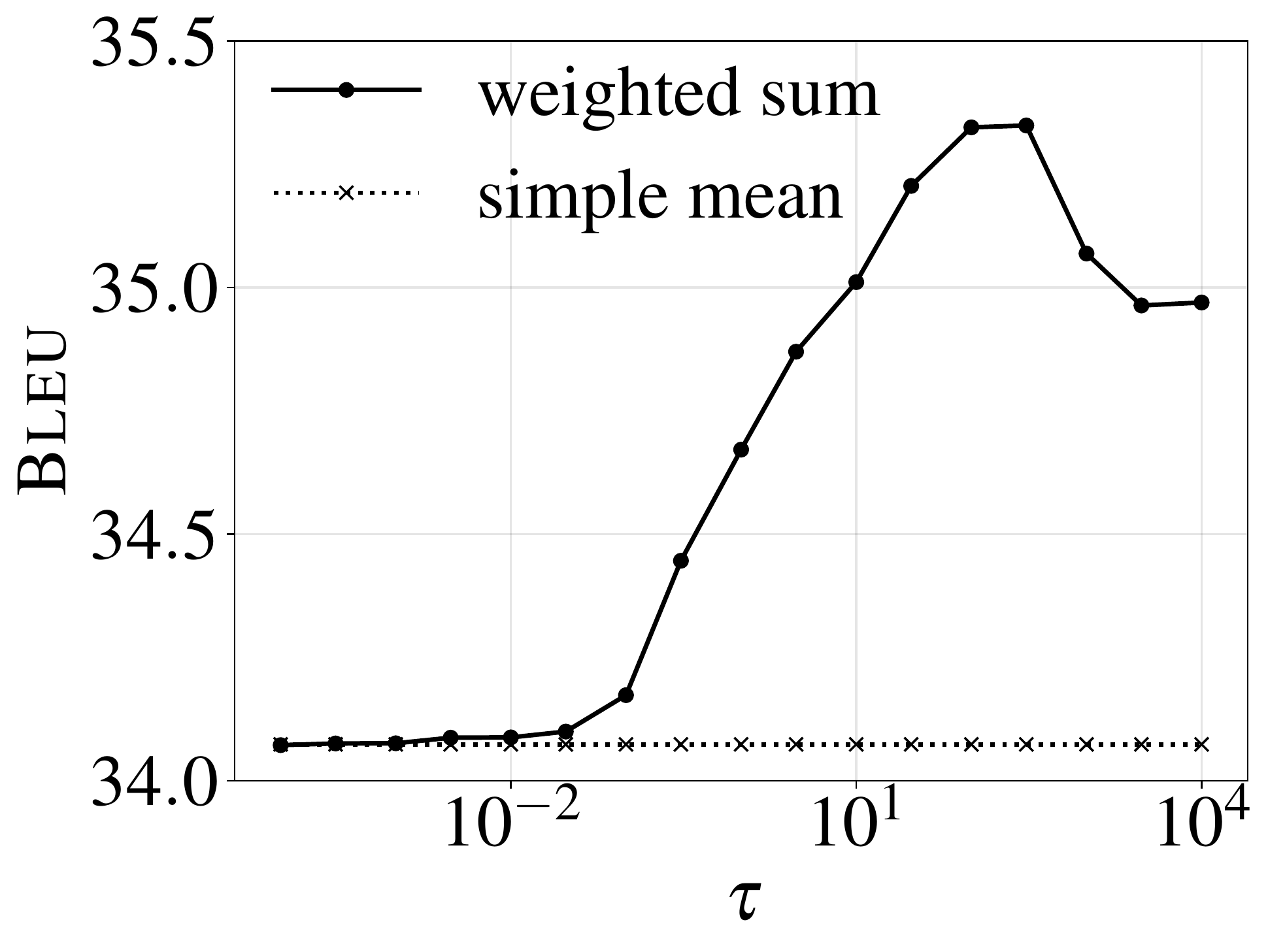}
    \caption{Last-$40$ weighted sum on de-en.}
    \label{fig.de-en.weighted-sum}
\end{figure}

In Fig.\ref{fig.de-en.weighted-sum}, we plot the \textsc{Bleu} scores against the temperature $\tau$ in Eq.\ref{eq.extd.}.
Here, we select last-$K$ checkpoints as in Fig.\ref{fig.de-en.lastK} to artificially include some bad-performing checkpoints.
Two sanity checks can be done here.
When $\tau$ is very small, uniform weights are used and the performance is close to the vanilla last-40 case.
When $\tau$ is very large, one-hot weights are used and the performance is close to that of the best checkpoint.
We observe that using the \textsc{DevPpl}-dependent weights results in similar performance increase compared to the vanilla case, meaning that the checkpoint selections can be automated by selecting a proper $\tau$.

\begin{figure}[ht]
    \centering
    \includegraphics[width=0.38\textwidth]{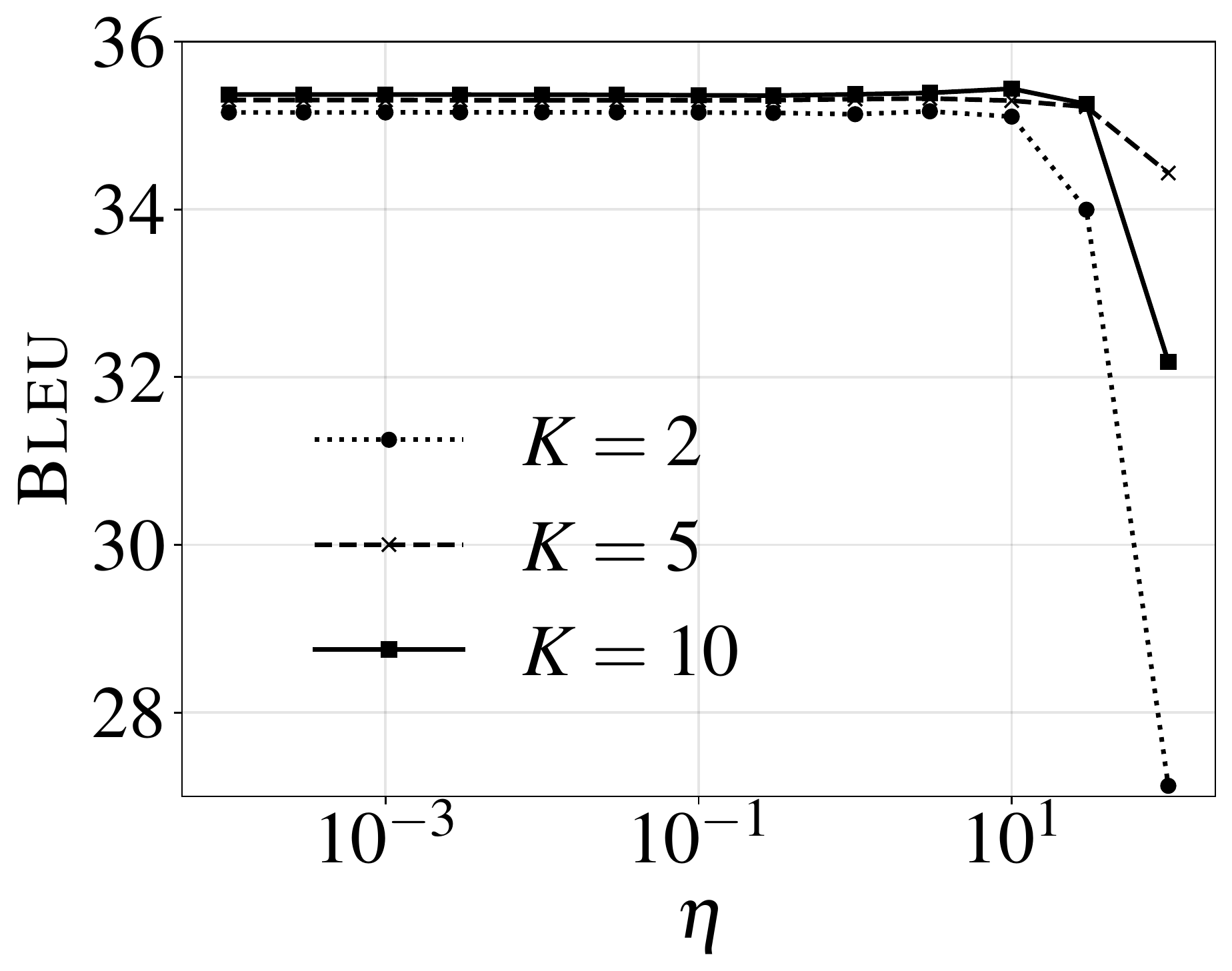}
    \caption{One-shot gradient update of top-$K$ weighted sum with $\tau=100$ on de-en.}
    \label{fig.de-en.oneshot-grad}
\end{figure}

Next, we study how the system performance changes with the step size used in the one-shot gradient update (Fig.\ref{fig:2} and Eq.\ref{eq.grad.}).
As shown in Fig.\ref{fig.de-en.oneshot-grad}, we interpolate three systems selecting top-$K$ checkpoints with $K=2$, $K=5$ and $K=10$, respectively.
Here, temperature $\tau=100$.
In line with the results in Fig.\ref{fig.de-en.lastK} and Fig.\ref{fig.de-en.topK}, the models with $K=5$ and $K=10$ are slightly better than the model with $K=2$.
However, as the step size $\eta$ increases, the \textsc{Bleu} score quickly drops as the averaged model diverges further away from the initial mean.
It is clear from the figure that nothing is gained in terms of \textsc{Bleu} during the $\eta$ scan.
In other words, these results suggest a very flat surface along the direction of averaged gradients.

\begin{figure}[ht]
    \centering
    \includegraphics[width=0.44\textwidth]{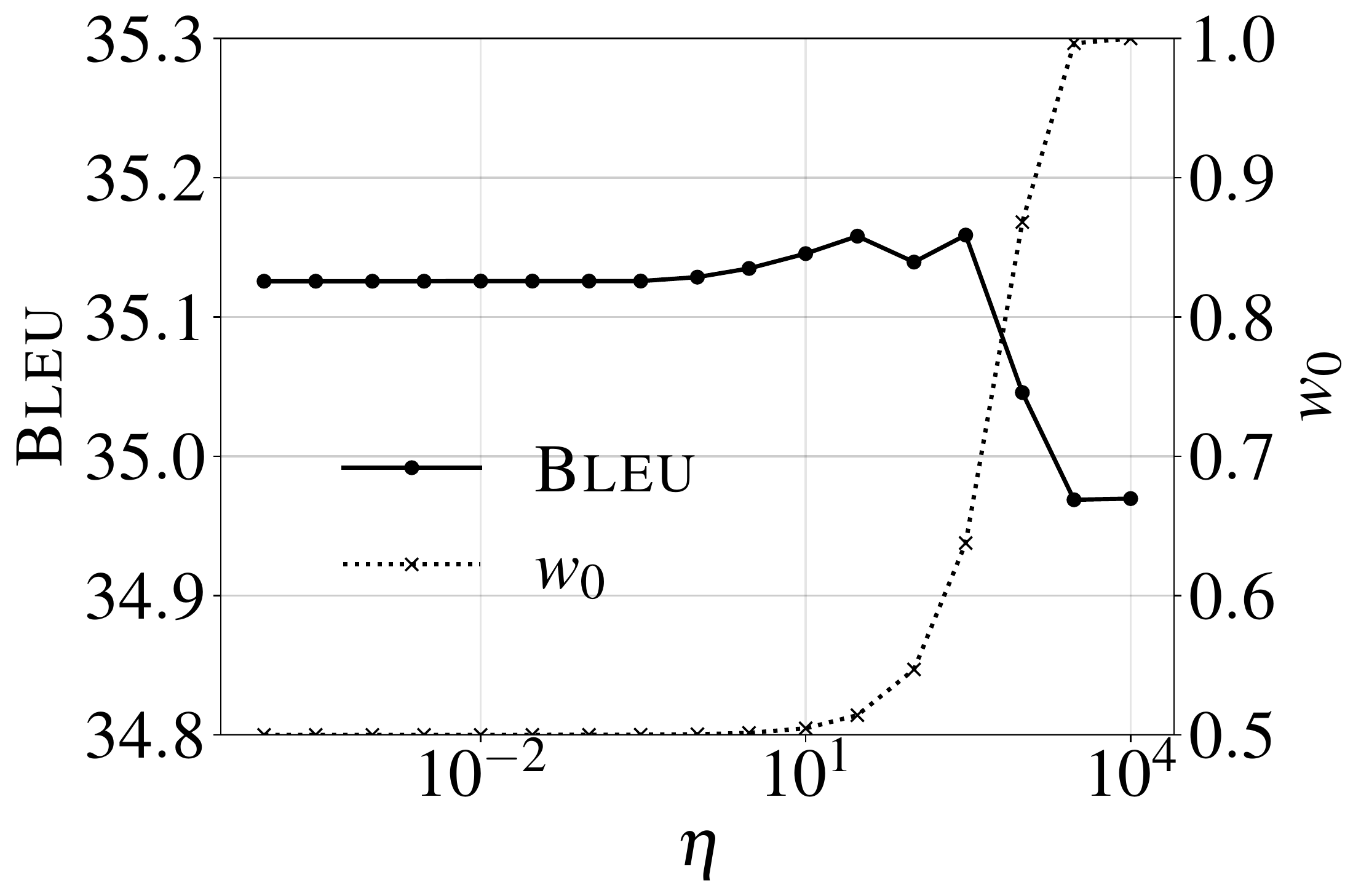}
    \caption{Optimization of interpolation weights $w_k$ on development data with $K=2$ on de-en.}
    \label{fig.de-en.optimize-on-valid}
\end{figure}

To investigate if optimization on the development data would work, we implement Eq.\ref{eq.opti.} and sweep over step size $\eta$.
As shown in Fig.\ref{fig.de-en.optimize-on-valid}, the gradient update on the weights move the model towards the best checkpoint ($\bm{\theta}_0$ here), and $w_0$ increases to 1.0 with large enough $\eta$.
There is, however, little improvement to be obtained along the path.
Note that this is the restricted case (Eq.\ref{eq.opti.}) where only interpolation weights are allowed to change and model parameters are not updated.

\begin{figure}[ht]
    \centering
    \includegraphics[trim={0 10em 0 3.6em},clip,width=0.44\textwidth]{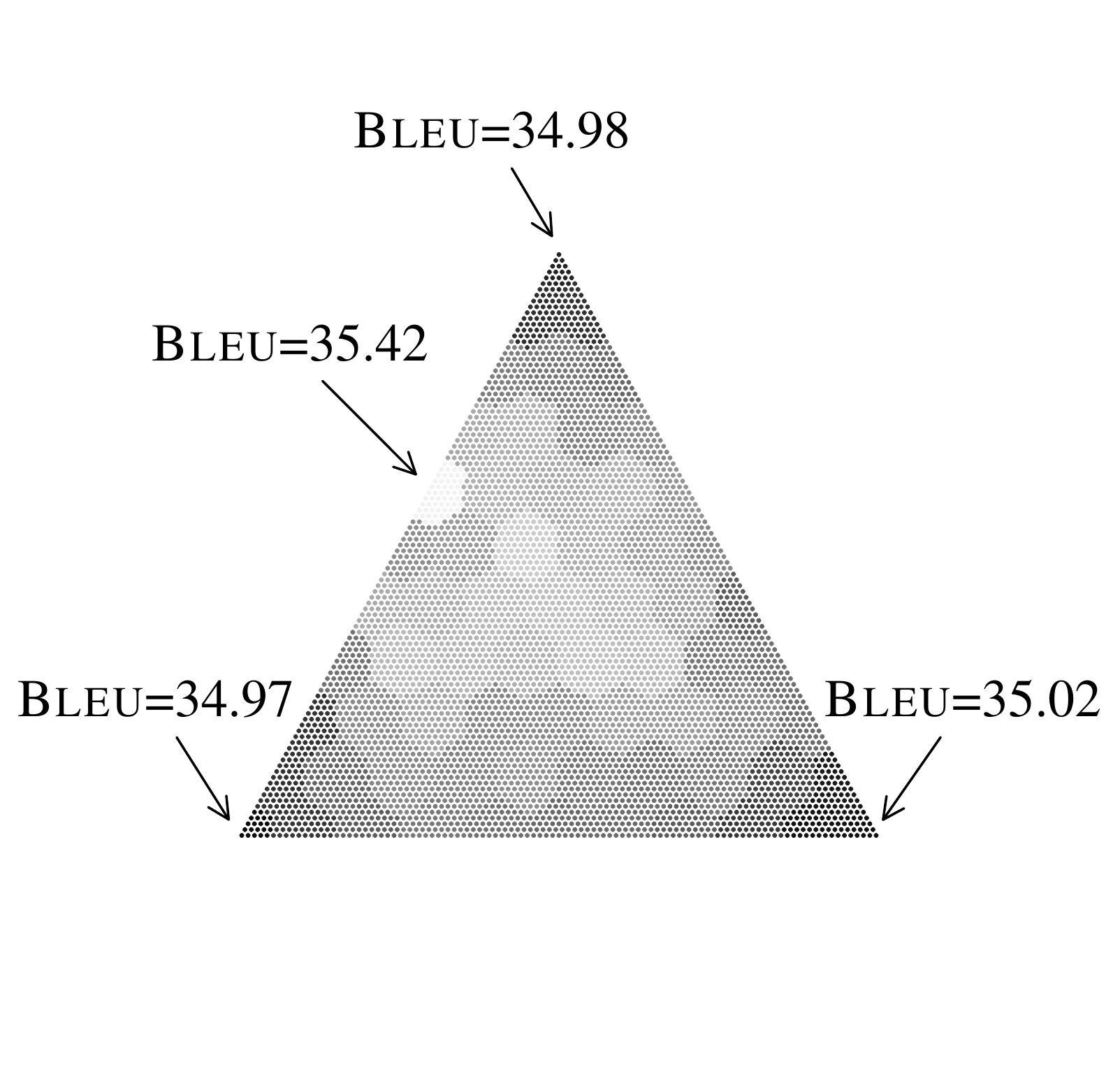}
    \caption{Neighborhood of the top-3 checkpoints on de-en. The hexagons are artifacts from plotting because a denser grid of points is used in the plot than in checkpoint averaging and the dots are colored by querying the nearest neighbor in the checkpoint averaging grid.}
    \label{fig.de-en.neighborhood}
\end{figure}

Given the results so far, it is clear that although a small boost of \textsc{Bleu} score can be robustly obtained in various checkpoint averaging settings, it is hard to squeeze out any further improvement with the extensions considered here.
We therefore perform a grid search over the interpolation weights $w_k$ with $K=3$, to examine the landscape between the checkpoints.
Shown in Fig.\ref{fig.de-en.neighborhood}, is the intersection of $w_1 + w_2 + w_3 = 1, 0 \leq w_k \leq 1$ in the space of the interpolation weights.
From the figure, except when really close to the vertices, i.e. $(w_1, w_2, w_3) = (1,0,0)$ or $(0,1,0)$ or $(0,0,1)$, the surface is rather flat with small fluctuations here and there.
Considered together with the previous results, this suggests that the gradient direction in the flat area may be unreliable and not much improvement is to be gained by further tuning the interpolation weights.
Of course one could argue that in higher dimensions the surface could look different by moving off of the $\sum_{k \in \mathcal{S}}w_k=1$ hyper-plain, but we think it is unlikely to be helpful as Fig.\ref{fig.de-en.oneshot-grad} is a counter-evidence at hand.

\vspace{-1em}
\section{Conclusion}

We consider checkpoint averaging, a simple and effective method in neural machine translation to boost system performance.
Specifically, we examine different checkpoint selection strategies, calculate weighted average, make use of gradient information and optimize the interpolation weights.
We confirm the robust improvements from checkpoint averaging and that the checkpoint selection can be automated with the weighted average scheme.
However, by closely looking at the landscape between the checkpoints, we find the surface to be rather flat and conclude that tuning in the space of the interpolation weights may not be a meaningful direction to squeeze out further improvements.

\section*{Acknowledgements}
This work was partially supported by the project HYKIST funded by the German Federal Ministry of Health on the basis of a decision of the German Federal Parliament (Bundestag) under funding ID ZMVI1-2520DAT04A, and by NeuroSys which, as part of the initiative “Clusters4Future”, is funded by the Federal Ministry of Education and Research BMBF (03ZU1106DA).

\bibliography{ref}
\bibliographystyle{acl}

\clearpage
\begin{appendices}
\section{Additional Results}

As mentioned, only results on de-en are reported in Sec.\ref{sec.exp.}.
In this section, further results on the other datasets are shown.

The data statistics are summarized in Tab.\ref{tab.stat.}.

\begin{table}[ht]
\centering
\begin{tabular}{crrr}
\hline
dataset & vocab & train pairs & test pairs \\ \hline
ru-en & 10k & 150k & 5.5k \\
de-en & 10k & 160k & 6.8k \\
es-en & 10k & 170k & 5.6k \\ \hline
en-ro & 20k & 0.6M & 2.0k \\
en-de & 44k & 4.0M & 3.0k \\
zh-en & 47k & 17.0M & 4.0k \\ \hline
\end{tabular}
\caption{\label{tab.stat.} Statistics of the datasets.}
\end{table}

\begin{figure}[ht]
    \centering
    \includegraphics[width=0.45\textwidth]{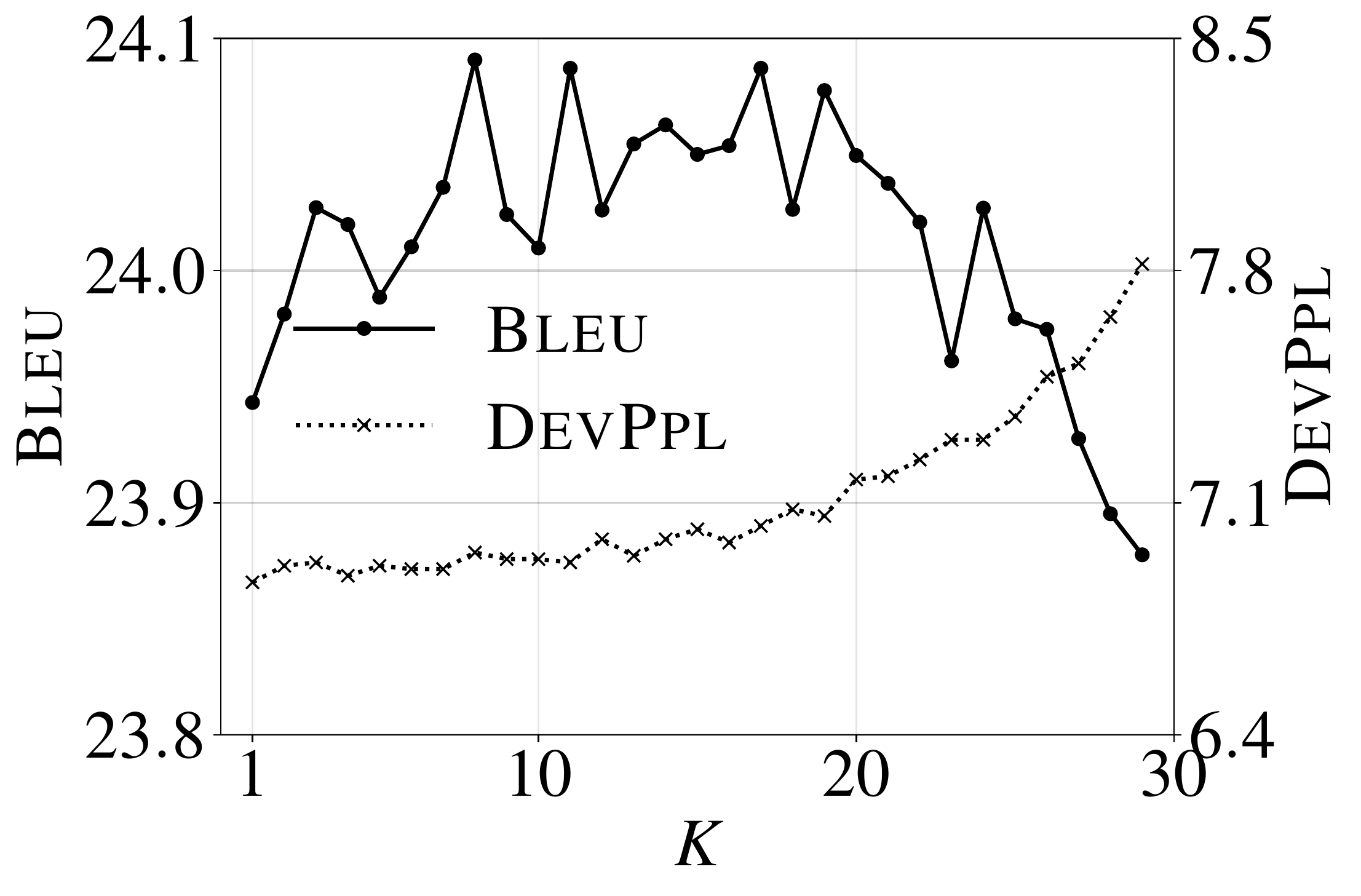}
    \caption{Last-$K$ simple mean on ru-en.}
    \label{fig.ru-en.lastK}
\end{figure}

Fig.\ref{fig.ru-en.lastK} shows the last-$K$ simple mean \textsc{Bleu} and \textsc{DevPpl} curves on ru-en.
As can be seen, the degredation of the interpolated models starts to happen when checkpoints with worse perplexities are included into the mixture.

\begin{figure}[ht]
    \centering
    \includegraphics[width=0.45\textwidth]{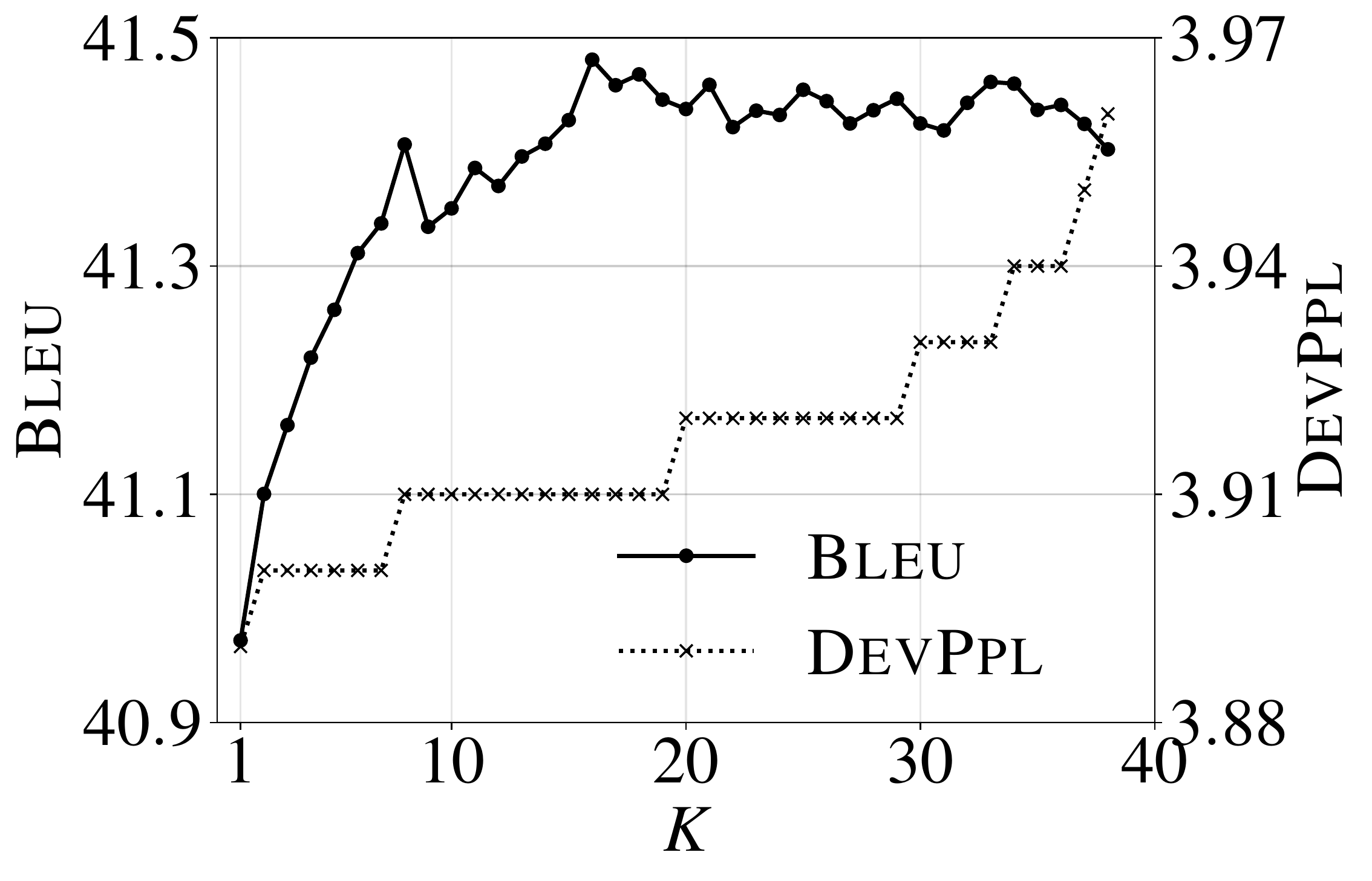}
    \caption{Top-$K$ simple mean on es-en.}
    \label{fig.es-en.topK}
\end{figure}

Fig.\ref{fig.es-en.topK} shows the top-$K$ simple mean \textsc{Bleu} and \textsc{DevPpl} curves on es-en.
Note that when all checkpoints are of decent \textsc{DevPpl}, the \textsc{Bleu} score of the averaged model is more stable.

\begin{figure}[ht]
    \centering
    \includegraphics[width=0.38\textwidth]{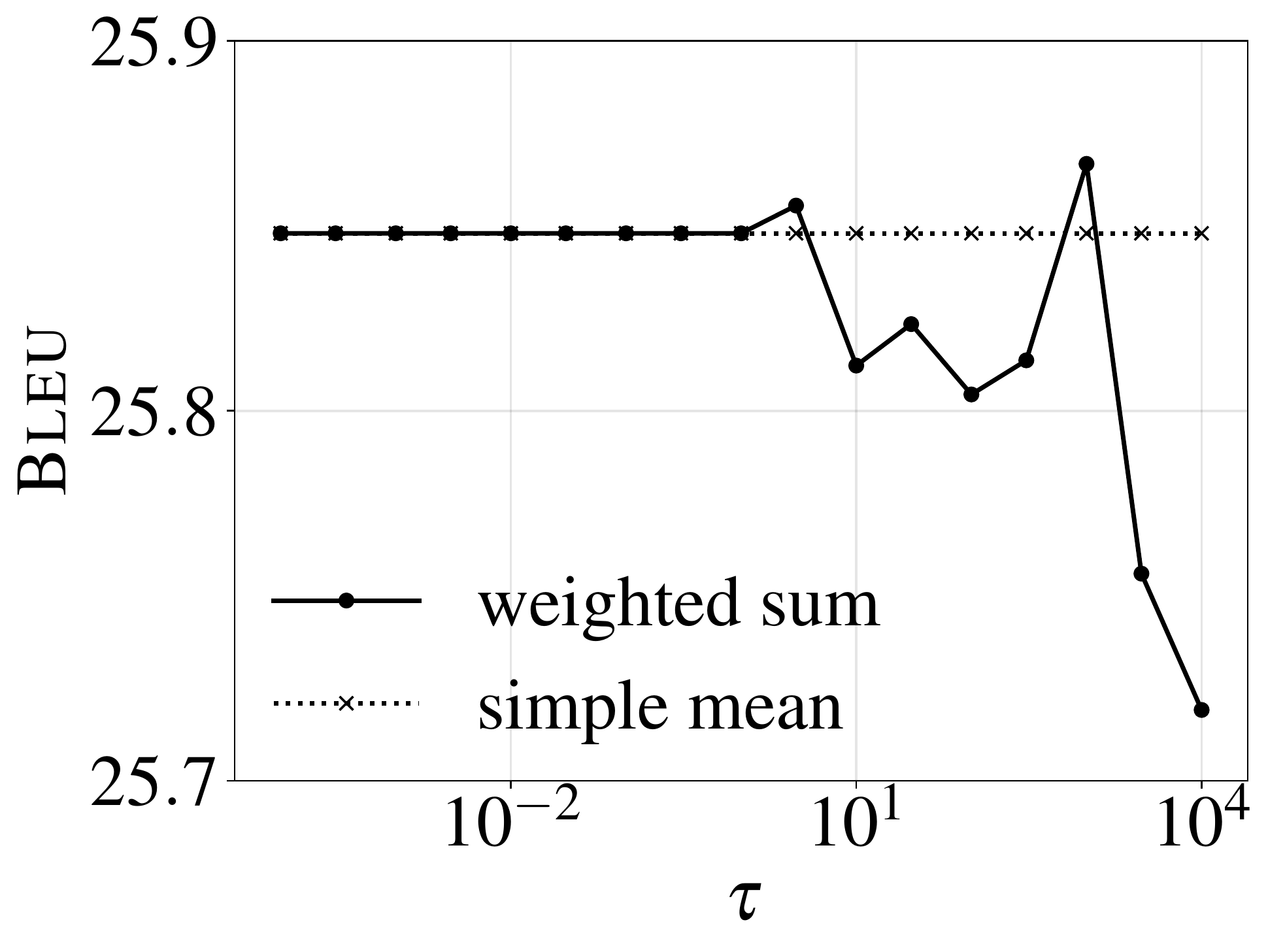}
    \caption{Top-10 weighted sum on en-ro.}
    \label{fig.en-ro.weighted-sum}
\end{figure}

Fig.\ref{fig.en-ro.weighted-sum} shows the top-10 weighted sum on en-ro.
Earlier in Fig.\ref{fig.de-en.weighted-sum}, we select last-40 checkpoints to include some bad-performing checkpoints.
Here, the top-10 checkpoints are selected and it is clear from the figure that there is not much to be gained when tuning the interpolation weight via the temperature hyperparameter $\tau$.

\begin{figure}[ht]
    \centering
    \includegraphics[trim={0 10em 0 3.6em},clip,width=0.44\textwidth]{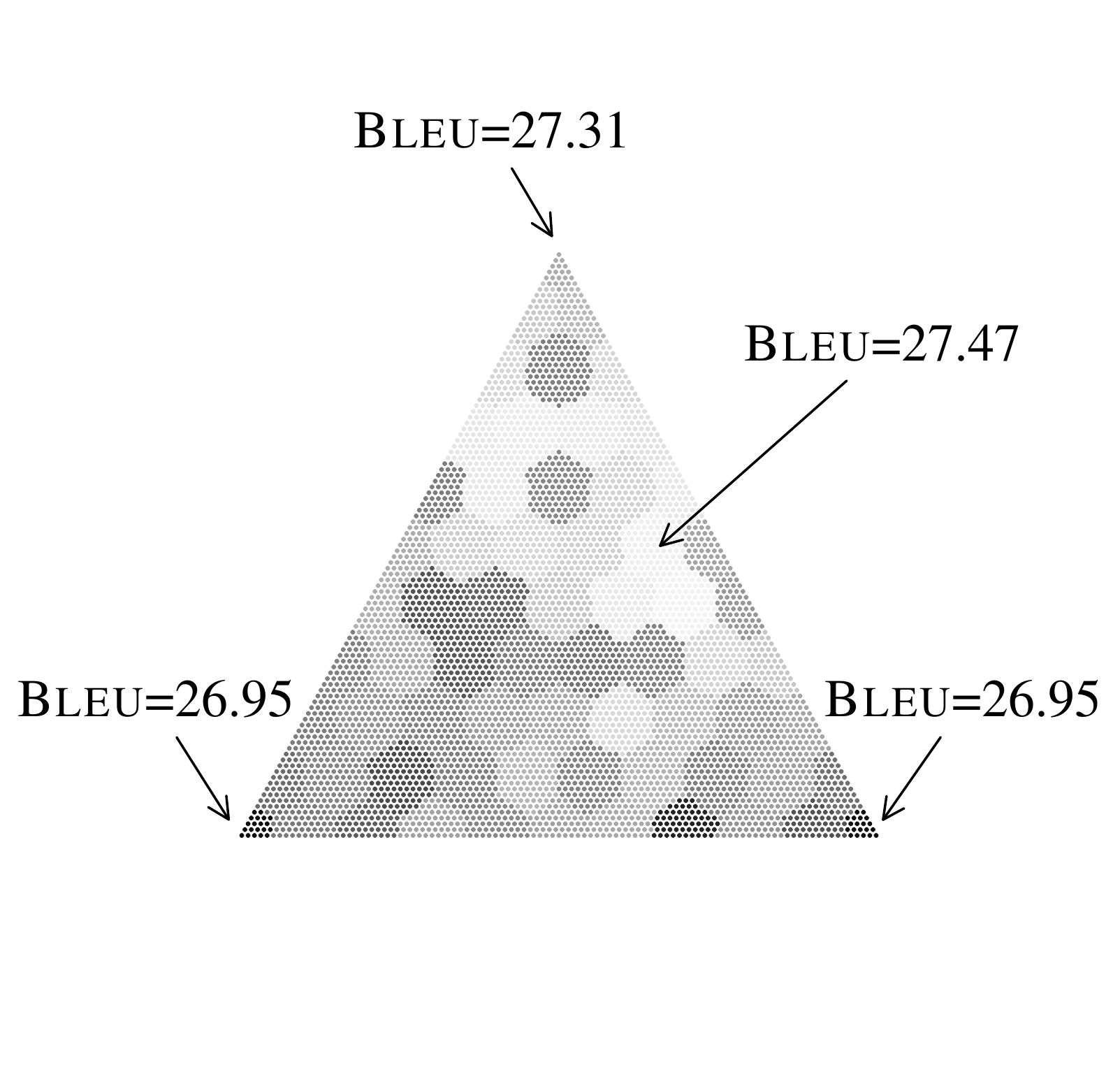}
    \caption{Neighborhood of three checkpoints on en-de. One good checkpoint and two relatively worse checkpoints are included to show the difference compared with Fig.\ref{fig.de-en.neighborhood}. No post-processing of splitting hyphenated compound words is done (See \url{https://github.com/tensorflow/tensor2tensor/blob/master/tensor2tensor/utils/get_ende_bleu.sh}.). The hexagons are artifacts from plotting because a denser grid of points is used in the plot than in checkpoint averaging and the dots are colored by querying the nearest neighbor in the checkpoint averaging grid.}
    \label{fig.en-de.neighborhood}
\end{figure}

In Fig.\ref{fig.en-de.neighborhood}, we plot the neighborhood of three checkpoints on en-de.
Here, One good checkpoint and two relatively worse checkpoints are included to show the difference compared with Fig.\ref{fig.de-en.neighborhood}.
As can be seen, the area near the good checkpoint is overall brighter and the region closer to the two worse checkpoints is darker.
Although noise is visible from the plot, it is clear that there is not a specific optima where the \textsc{Bleu} score of the checkpoint-averaged model is significantly better.

\begin{figure}[ht]
    \centering
    \includegraphics[trim={0 10em 0 3.6em},clip,width=0.44\textwidth]{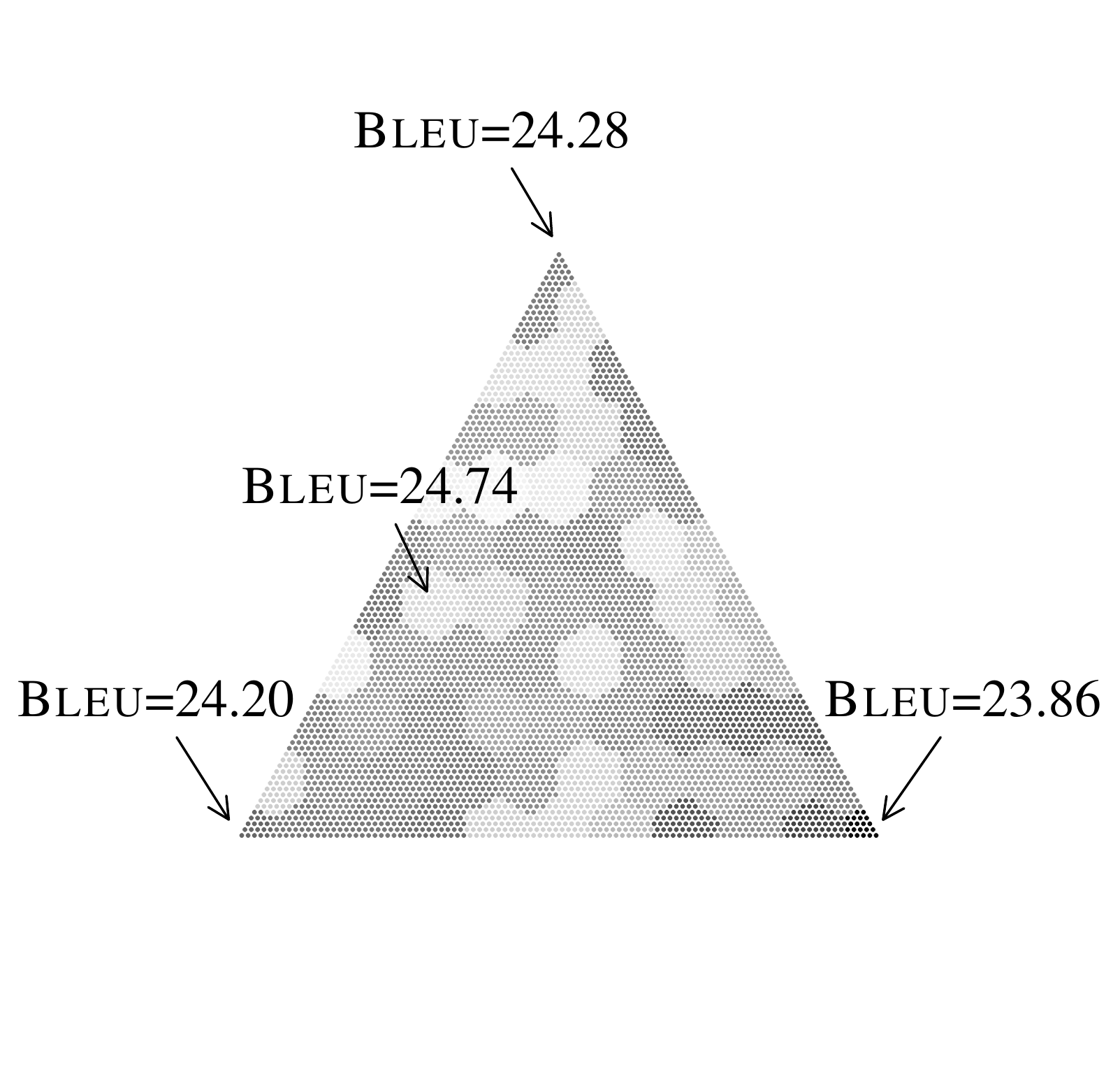}
    \caption{Neighborhood of three checkpoints on zh-en. Two good checkpoint and one relatively worse checkpoint are included to show the difference compared with Fig.\ref{fig.de-en.neighborhood}. The hexagons are artifacts from plotting because a denser grid of points is used in the plot than in checkpoint averaging and the dots are colored by querying the nearest neighbor in the checkpoint averaging grid.}
    \label{fig.zh-en.neighborhood}
\end{figure}

In Fig.\ref{fig.zh-en.neighborhood}, we further plot the neighborhood of three checkpoints on zh-en.
Here, two good checkpoint and one relatively worse checkpoint are included to show the difference compared with Fig.\ref{fig.de-en.neighborhood}.
From the figure, it can be seen that, overall, the interpolation closer to the two good checkpoints is better than when the worse checkpoint has a larger weight.
Although +0.4\% absolute \textsc{Bleu} score improvement is possible, there is no further improvement to be gained when tuning the interpolation weights.

\end{appendices}

\end{document}